\documentclass[letterpaper, 10 pt, conference]{ieeeconf}  

\IEEEoverridecommandlockouts                              

\title{\LARGE \bf
Scaling Local Control to Large-Scale Topological Navigation}

\author{Xiangyun Meng, Nathan Ratliff, Yu Xiang and Dieter Fox
\thanks{Xiangyun Meng and Dieter Fox are with the Paul G. Allen School of Computer Science \& Engineering, University of Washington, Seattle, WA 98195, USA {\tt\small\{xiangyun, fox\}@cs.washington.edu}}
\thanks{Nathan Ratliff, Yu Xiang and Dieter Fox are with NVIDIA, Seattle, WA 98105, USA {\tt\small \{ nratliff, yux, dieterf\}@nvidia.com }}
}

\usepackage{hyperref}
\usepackage{xcolor}
\usepackage{amsmath}
\usepackage{amssymb}
\usepackage{mathtools}
\usepackage{siunitx}
\usepackage{graphicx}
\usepackage{adjustbox}
\usepackage{float}
\usepackage{wrapfig}
\usepackage[labelformat=simple,font=footnotesize]{subcaption}
\usepackage[font=small,skip=3pt,tableposition=top]{caption}
\usepackage{booktabs}

\newcommand{\argmax}{\mathop{\mathrm{argmax}}}
\newcommand{\C}{\mathbb{C}}
\newcommand{\RE}{\mathbb{RE}}
\def\figurespace{\vspace{-3.2ex}}

\begin{document}

\IEEEaftertitletext{\vspace{-1.5\baselineskip}}
\maketitle

\thispagestyle{empty}
\pagestyle{empty}

\begin{abstract}
Visual topological navigation has been revitalized recently thanks to the advancement of deep learning that substantially improves robot perception.  However, the scalability and reliability issue remain challenging due to the complexity and ambiguity of real world images and mechanical constraints of real robots. We present an intuitive approach to show that by accurately measuring the capability of a local controller, large-scale visual topological navigation can be achieved while being scalable and robust. Our approach achieves state-of-the-art results in trajectory following and planning in large-scale environments. It also generalizes well to real robots and new environments without retraining or finetuning. 
\end{abstract}

\section{Introduction}


There has been an emergence of cognitive approaches \cite{savinov2018semi, wubayesian, kumar2018visual,blochliger2018topomap, chen2019graphnav} towards navigation thanks to the advancement of deep learning that substantially improves robot perception. Compared to the traditional mapping, localization and planning  approach (SLAM) \cite{thrun2005probabilistic, mur2015orb} that builds a metric map, cognitive navigation uses a topological map. This eliminates the need of meticulously reconstructing an environment which requires expensive or bulky hardware such as a laser scanner or a high-resolution camera. Moreover, the fact that humans are able to navigate effortlessly in large-scale environments without a metric map is intriguing. By adding this cognitive spatial reasoning capability to robots, we could potentially lower the hardware cost (i.e., using low-resolution cameras), make them work more robustly in dynamic environments and bring insights to more complex tasks such as visual manipulation.

While cognitive navigation has drawn significant attention recently, the problem remains challenging because i) it does not scale well to the size of experiences ii) it is fragile due to actuation noise and dynamic obstacles and iii) it lacks probabilistic interpretation, making it difficult to plan with uncertainty. These problems are exacerbated when using a RGB camera in indoor environments, where partial observability makes it difficult to control a robot to follow a single path \cite{kumar2018visual, hirose2019deep}.
\begin{figure}
    \centering
    \includegraphics[width=0.9\columnwidth]{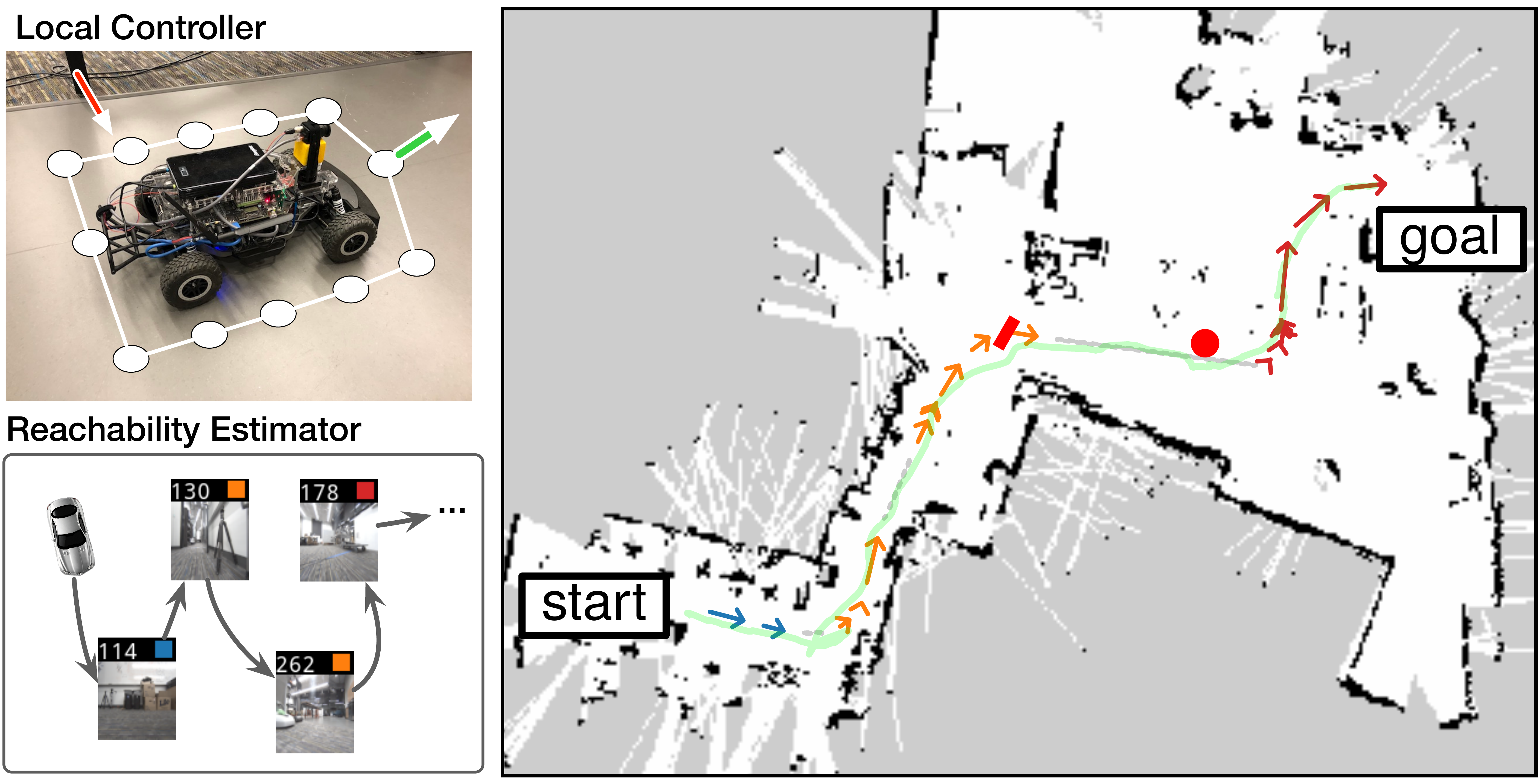}
    \caption{Overview of our method. The local controller drives the vehicle towards a given target image, and the reachability estimator plans a path by combining multiple experiences (colored arrows on the map) to provide the controller a sequence of target observations (bottom left) to follow. The vehicle is able to navigate robustly in the real environment (right) while avoiding unseen obstacles (red rectangle and circle). The model is trained entirely in simulation.}
    \label{fig:intro}
    \figurespace
\end{figure}

In this paper, we present a simple and intuitive solution for topological navigation. We show that by accurately measuring the capability of a local controller, robust visual topological navigation can be achieved with sparse experiences (Fig.\ref{fig:intro}). In our approach, we do not assume the availability of a global coordinate system or robot poses, nor do we assume noise-free actuation or static environment. This minimalistic representation only has two components: a local controller and a reachability estimator. The controller is responsible for local reactive navigation, whereas the reachability estimator measures the {\it capability} of the controller for landmark selection and long-term probabilistic planning. To achieve this, we leverage the Riemannian Motion Policy (RMP) framework \cite{rmp} for robust reactive control and deep learning for learning the capability of the controller from data. We show that with both components working in synergy, a robot can i) navigate robustly with the presence of nonholonomic constraints, actuation noise and obstacles; ii) build a compact spatial memory through adaptive experience sparsification and iii) plan in the topological space probabilistically, allowing robot to generalize to new navigation tasks.

We evaluate our approach in the Gibson simulation environment \cite{xia2018gibson} and on a real RC car. Our test environments contain a diverse set of real-world indoor scenes with presence of strong symmetry and tight spaces. We show that our approach generalizes well to these unseen environments and surprisingly well to real robots without finetuning. Scalability-wise, our spatial memory grows only when new experiences are unseen, making the system space-efficient and compute-efficient.

\begin{figure*}
    \centering
    \includegraphics[width=0.85\textwidth]{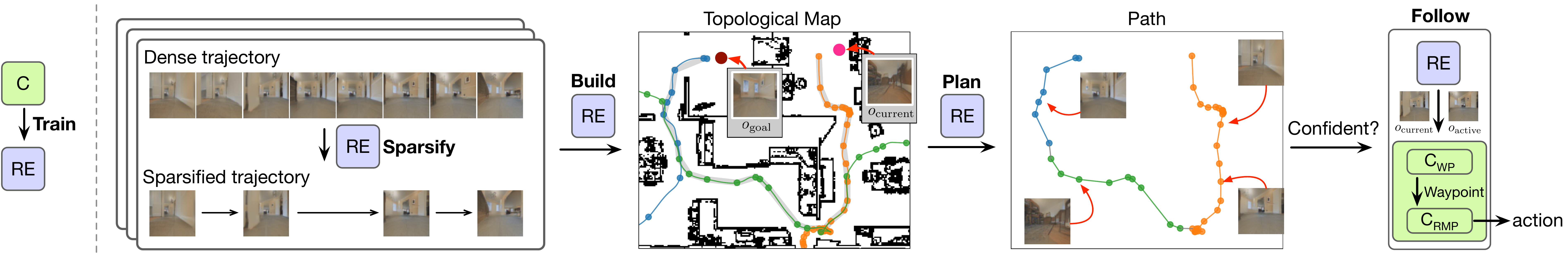}
    \caption{System overview. Given a controller $\C$, we train a reachability estimator $\RE$. $\RE$ is used for sparsifying incoming trajectories, building a compact topological map and planning a path. $\C$ and $\RE$ work in synergy to robustly follow the planned path.}
    \label{fig:system_overview}
    \figurespace
    \vspace{-0.5ex}
\end{figure*}

\section{Related Work}
Cognitive spatial reasoning has been extensively studied both in neuroscience \cite{o1978hippocampus,hafting2005microstructure, doeller2010evidence}, and robotics \cite{thrun1998learning, milford2004ratslam, kuipers2000spatial}. The Spatial Semantic Hierarchy \cite{kuipers2000spatial} divides the cognitive mapping process into four levels: control, causal, topological and metric. In our method, the local controller operates on the control level, whereas the reachability estimator reasons about causal and topological relationship between observations. We omit metric-level reasoning since we are not concerned about building a metric map.

Experience-driven navigation constructs a topological map for localization and mapping \cite{milford2004ratslam, maddern2012capping, fraundorfer2007topological, cummins2011appearance, blochliger2018topomap}. Unlike SLAM that assumes a static environment, the experience graph can also be used for dealing with long-term appearance changes \cite{linegar2015work}. This line of works mostly focus on appearance-based localization and ignore the control aspect of navigation, and assume that a robot can always follow experiences robustly. This does not usually hold in unstructured indoor environments, where it is crucial to design a good controller while considering its capability.

Semi-Parametric Topological Memory (SPTM) \cite{savinov2018semi, Savinov2019_EC} is a recent work that adopts deep learning into topological navigation. Similar to SPTM, we build a topological map through past experiences. Unlike SPTM that uses image similarity as a proxy for reachability, we measure the reachability of a controller directly. This significantly improves robustness and opens opportunities for constructing sparse maps.

There have been recent works studying visual trajectory following that handles obstacles \cite{hirose2019deep, bansal2019combining}, actuation noise \cite{kumar2018visual}, or with self-supervision \cite{pathak2018zero}. Our approach differs from them in that our trajectory follower extends seamlessly to probabilistic planning. Our method also handles obstacles and actuation noise well, thanks to the RMP controller that models local geometry and vehicle dynamics.

Recent works on cognitive planning \cite{gupta2017cognitive, gupta2017unifying} show that a neural planner can be learned from data. However, assumptions such as groundtruth camera poses are available with perfect self-localization are unrealistic. The use of grid map also limits its flexibility. Another line of research uses reinforcement learning to learn a latent map \cite{mirowski2018learning, mirowski2016learning}, but it is data-inefficient and cannot be easily applied to real robots. In contrast, our planner is general and can adapt to new environments quickly. It bears a resemblance to feedback motion planning system such as LQR-Trees \cite{tedrake2010lqr}, where planning is performed on the topological map connecting reachable state spaces with visual feedback control.


\section{Method}

\subsection{Overview}
We consider the goal-directed navigation problem: a robot is asked to navigate to a goal $G$ given an observation $o_G$ taken at $G$. Robot does not have a map of the environment, but we assume it has collected a set of trajectories (e.g., via self-exploration or following language instructions) as its experiences. Each trajectory is a dense sequence of observations $o_1, o_2, ..., o_N$ recorded by its on-board camera. Using its experiences, robot decides the next action to take in order to reach $G$. The action space is continuous (e.g., velocity and steering angle) and robot could be affected by actuation noise and unseen obstacles.

We approach this problem from a cognitive perspective. Robot first builds a topological map from its experiences. The map is a directed graph, with vertices as observations and edges encoding traversability. Then, given its current observation $o_\text{current}$ and goal $o_G$, robot searches for a path on the graph and follows that path to reach $G$. Our setup is similar to that of SPTM \cite{savinov2018semi}. The difference is that we design our system to make it generalize to real robots and scale to real environments.

For such a navigation system to work, we first need a target-conditioned \textbf{Local Controller} $\C$. $\C$ takes current observation and a target observation, and outputs an action $a=\C(o_\text{current}, o_{\text{target}})$ to drive robot towards the target. The action is executed for a small time step to get an updated $o_\text{current}$ and the process is repeated until $o_\text{current}$ matches $o_\text{target}$. Given a path (a sequence of observations) computed by a planner, robot uses $\C$ to follow the path progressively to reach its final destination.

In practice, robot's experience pool can be large and grow indefinitely, thus the key issue is to build a sparse and scalable representation of an environment given dense, unstructured trajectories. Clearly, adjacent observations in a trajectory is highly correlated and it would be wasteful to keep every observation. One ad-hoc approach to sparsify a trajectory is to take every $n$th observation. However, this assumes that target $n$ steps away is always reachable, which is not necessarily true. For example, without occlusion, an observation far away can be confidently reached (e.g., in a straight hallway), whereas an observation nearby may be hidden (thus not reachable) if it is blocked by obstacles.  Moreover, motion constraints, sensor field of view, motor noise, etc. can all affect the reachability of a target. 

Our intuition is that the sparsification of a trajectory should adapt to the capability of the controller. 
We propose learning a \textbf{Reachability Estimator} $\RE$ that predicts the probability of  $\C$ successfully reaching a target: $\RE(o_{\text{current}}, o_{\text{target}})=P(\text{reach} | o_{\text{current}}, o_{\text{target}}, \C)$. We use $\RE$ as a probabilistic metric throughout the system, illustrated by Fig.~\ref{fig:system_overview}. Given a controller $\C$, we train a corresponding $\RE$. The incoming trajectories are first sparsified by $\RE$ and then interlinked to form a compact topological map. Given $o_{\text{current}}$ and $o_{G}$, we leverage $\RE$ to plan a probabilistic path and use $\C$ and $\RE$ in synergy to follow the planned path robustly.

\subsection{Designing a Robust Local Controller}
\label{sec:controller}
Real-world robots are subject to disturbances such as motor noise and moving obstacles, which can cause a robot to deviate from planned path and fail. Hence our first objective is to design a sufficiently robust local controller. Contrary to directly predicting low-level controls, we split our controller into two stages: high-level waypoint prediction and low-level reactive control. The high-level controller $\C_{\text{WP}}$ predicts a waypoint $x, y$ (in robot's local coordinate system) for the low-level controller. The waypoint needs not be precise, but only serves as a hint for the low-level controller to make progress. Hence $\C_{\text{WP}}$ is agnostic to robot dynamics (e.g., can be trained with A* waypoints as supervision) and absorbs the effects of actuation noise. For the low-level controller, we adopt the RMP representation \cite{meng2019neural} as a principled way for obstacle avoidance and vehicle control. Hence we have $\C(o_{\text{current}}, o_{\text{target}})=\C_{\text{RMP}}(\C_{\text{WP}}(o_\text{current}, o_{\text{target}}))$. Note that this allows the same $\C_{\text{WP}}$ to be applied to different robots by replacing the low-level controller.

\begin{figure}
    \centering
    \includegraphics[width=0.92\columnwidth]{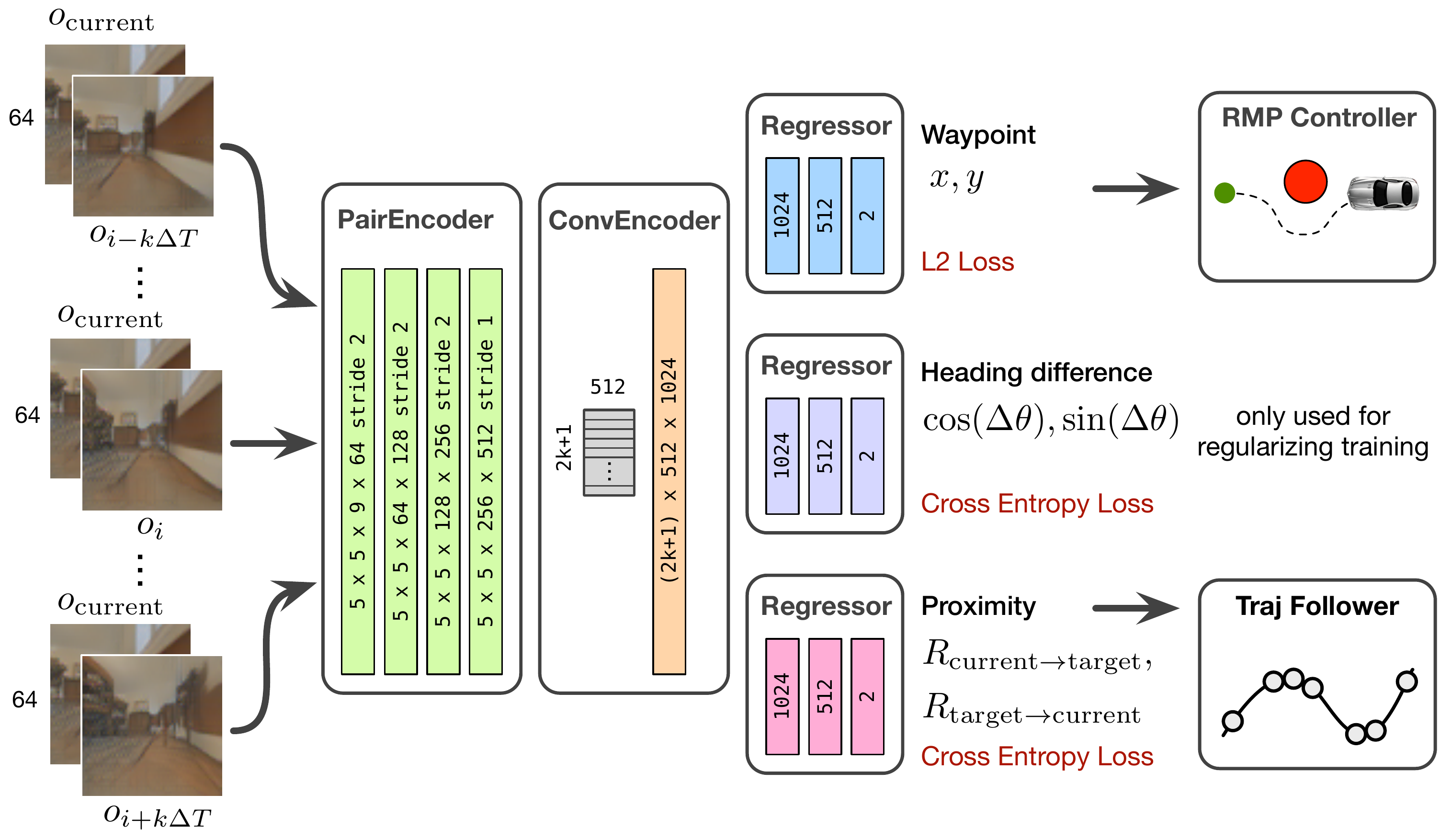}
    \caption{Architecture of $\C_{\text{WP}}$. The architecture of $\RE$ is similar, except that it regresses to a single probability and is supervised with cross-entropy loss.}
    \label{fig:network}
    \figurespace
\end{figure}

Fig.~\ref{fig:network} illustrates the design of $\C_{\text{WP}}$. The robot state is represented by its current observation $o_{\text{current}}$. Denote $i$th observation in a trajectory as $o_i$. We represent the corresponding $o_{\text{target}}$ at $o_{i}$ as a sequence of neighbor observations centered at $o_i$: $$o_{i-k\Delta T}, o_{i-(k-1)\Delta T}, ..., o_i,..., o_{i+(k-1)\Delta T}, o_{i+k\Delta T},$$ where $k$ controls  context length and $\Delta T$ (set to 3) is the gap between two observations. The past frames expand the field of view of $o_i$ which helps controller to do visual closed-loop control. The future frames encode intention at $o_i$, allowing a controller to adjust its waypoint in advance in order to follow subsequent targets smoothly and reliably.

Technically, we extract a feature vector by feeding stacked $[o_{\text{current}}, o_{i-k\Delta T}, o_{\text{current}} - o_{i-k\Delta T}]$  into a sequence of convolutions, followed by combining the $2k + 1$ feature vectors through one convolution and multiple fully-connected layers to predict a waypoint $x, y$. We find this design works much better than featurizing each image or stacking all images together. Additionally, the network predicts the heading difference between $o_{\text{current}}$ and $o_i$ to help the network anchor the target image in the sequence. Finally, in order to reason about proximity to a target (Sec.\ref{sec:follow_traj}), $\C_\text{WP}$ predicts mutual image overlap. Image overlap is a ratio that represents the percentage of content in one image that is visible in another image. Hence mutual image overlap is a pair of ratios $(R_{\text{current}\rightarrow \text{target}}, R_{\text{target}\rightarrow \text{current}})$. 


\begin{table}
  \centering
  \footnotesize
  \begin{tabular}{l|*{5}{c}}
    \toprule
    & Control(k=0) & Ours(k=0) & Ours(k=2) & Ours(k=5)\\
    \hline
    Success\% & 46\% & 88\% & 91\% & 95\%\\
    \bottomrule
  \end{tabular}
  \caption{Success rate for each controller.}
  \label{tab:controller_success_rate}
  \figurespace
  \vspace{-3ex}
\end{table}

We train $\C_{\text{WP}}$ in a supervised fashion (Sec.\ref{sec:experiments}). To evaluate our design, we randomly sample $o_{\text{current}}$ from a trajectory and $o_{\text{target}}$ being $-1.0m$ behind to $3.0m$ ahead of $o_{\text{current}}$ in 4 unseen environments, and run each controller to see if robot reaches the target. Table~\ref{tab:controller_success_rate} compares the success rate of each controller. Directly predicting low-level controls (forward acceleration and steering velocity) results in much lower success rate than our two-stage design. Compared with directly mapping images to low-level actions, we find it more robust to map images to higher-level abstractions such as waypoints, and then map waypoints to low-level controls using a representation (e.g., RMP) that explicitly models environmental geometry and robot dynamics.




\subsection{Learning the Reachability Estimator}
Table~\ref{tab:controller_success_rate} suggests that controller design and parametrization can heavily affect target reachability. Unlike \cite{savinov2018semi} that uses image similarity as a proxy, we learn reachability by explicitly predicting the execution outcome of $\C$. During training, $o_{\text{current}}$ and $o_{\text{target}}$ are randomly sampled from demonstration trajectories (Sec.~\ref{sec:experiments}) and $\C$ is used to drive the robot from $o_{\text{current}}$ to $o_{\text{target}}$ to get a binary outcome. The criteria for success is that robot reaches the target within time limit defined as $t_\text{max} = \text{A*}(o_\text{current}, o_\text{target}) / v_\text{min}$, where $\text{A*}(\cdot, \cdot)$ computes the $\text{A*}$ path length and $v_\text{min}$ is the minimum velocity. Hence $\RE$ measures the probability of $\C$  reaching the target \emph{efficiently}, which is independent of the temporal and physical distance between $o_\text{current}$ and $o_\text{target}$. This idea has an interesting connection to feedback motion planning systems \cite{tedrake2010lqr}, as $\RE$ can be seen as estimating visual funnels that are locally stable.


The design of $\RE$ is almost identical to $\C$, except that it predicts a single probability and is trained with a binary classification loss.

\label{sec:planning}


\subsection{Sparsifying a Trajectory}
For any observation $o_i$ in a dense trajectory, if $\RE(o_i, o_{i+1}), ..., \RE(o_i, o_{i+k+1})$ are sufficiently high, we could confidently discard $o_{i+1}, ..., o_{i+k}$ because $\C$ does not need them to reach $o_{i+k+1}$. Hence a greedy approach to choose the next anchor is
\begin{align*}
    \max & \ j \\
    \text{s.t. } & \RE(o_{i}, o_k) > p_{\text{sparsify}}, \forall k, i < k \leq j
\end{align*}
where $i$ is previous anchor's position and $p_{\text{sparsify}}$ is the probability threshold that controls sparsity. Hence a dense trajectory is converted to a sequence of contextified anchor observations $\hat{o}_1, ..., \hat{o}_m$. One may argue that contextification reduces the effective sparsification ratio. Since the time and space complexity is a function of the number of anchors, in practice it significantly saves computation during planning and following a trajectory, allowing our system to run on a robot in real time.


\subsection{Building a Compact Probabilistic Topological Map}
\begin{figure}
    \centering
    \includegraphics[width=0.8\columnwidth]{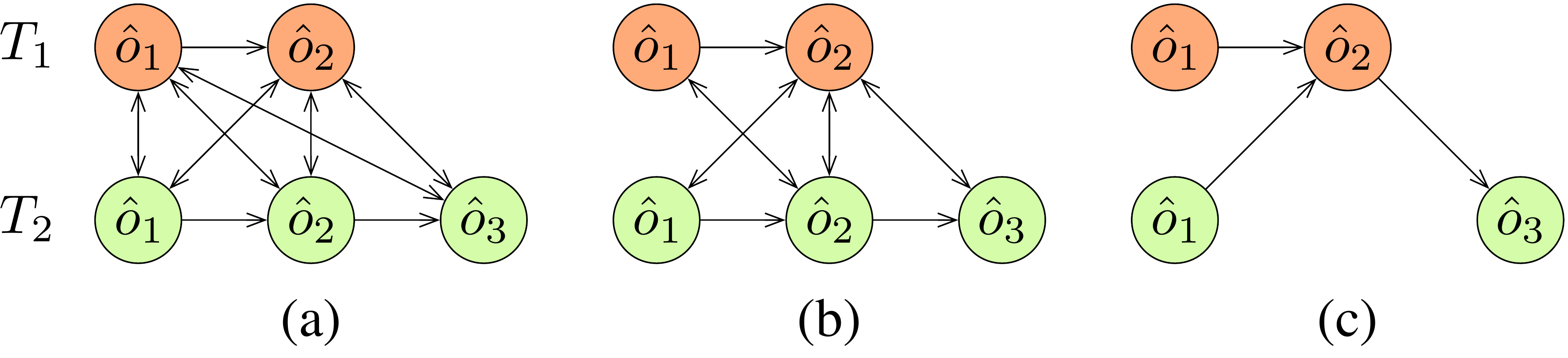}
    \vspace{-1ex}
    \caption{A topological map containing two trajectories. (a) densely connected graph. (b) after pruning low-probability edges. (c) after reusing nodes.}
    \label{fig:topomap}
    \figurespace
    \vspace{-1ex}
\end{figure}

Our topological map is a weighted directed graph (Fig.~\ref{fig:topomap}a). Vertices are anchor observations and edge weight from $\hat{o_i}$ to $\hat{o}_j$ is $-\log \RE(\hat{o_i}, \hat{o_j})$. Construction is incremental: for an incoming trajectory, we create pairwise edges between every vertex in the graph and every anchor in the new trajectory.

Compared to a graph constructed with dense observations, a graph built from sparsified observations has less than 1/10 of the vertices and 1/100 of the edges. To further improve scalability, we propose the following two optimizations to make the graph grow sublinearly to the number of raw observations, and eventually the size of the graph converges:

{\bf Edge pruning}. Low-probability edges with $\RE(\hat{o_i}, \hat{o_j}) < p_{\text{edge}}$ are discarded since they contribute little to successful navigation (Fig.~\ref{fig:topomap}b).

{\bf Vertex reuse}. It is common for two trajectories to be partially overlapped and storing this overlapping part repeatedly is unnecessary. Hence when adding anchor $\hat{o}_i$ into a graph, we check if there exists a vertex $\hat{o}$ such that $\RE(\hat{o}_{i-1}, \hat{o}) > p_{\text{reuse}}$ and $\RE(\hat{o}, \hat{o}_{i+1}) > p_{\text{reuse}}$. If the condition holds, we discard $\hat{o}_i$ and add edges $\hat{o}_{i-1} \rightarrow \hat{o}$ and $\hat{o} \rightarrow \hat{o}_{i+1}$, as illustrated in Fig.~\ref{fig:topomap}c.

The graph will converge because for any static environment of finite size, there is a maximum density of anchors. Any additional anchor will pass the vertex reuse check and be discarded. Practically however, an environment may change over time. The solution is to timestamp every observation and discard outdated observations using $\RE$. We leave the handling of long-term appearance change as future work.

\subsection{Planning}
We add an edge (weighted by its negative log probability) from $o_{\text{current}}$ to every vertex in the graph, and from every vertex in the graph to $o_G$. The weighted Dijkstra algorithm computes the path with the lowest negative log probability (i.e., the path that robot is most confident). Robot then decides whether the probability is high enough and may run the trajectory follower proposed in Section~\ref{sec:follow_traj}.


\subsection{Mitigating Perceptual Aliasing}
\label{sec:mitigate_aliasing}
Practically, $o_{\text{current}}$ may correspond to different locations of similar appearances. Traditional approaches usually formulate this as a POMDP problem \cite{thrun2005probabilistic} and try to resolve the ambiguity by maintaining beliefs over states. This requires having a unique state (e.g., global pose) associated with each observation which is difficult to implement since we do not have any metric information.

We use two techniques to resolve ambiguity. The first is to match a sequence of anchors during search and graph construction. In practice the probability of two segments having similar appearances is much lower than two single observations. Additionally we let robot re-plan a new path if it detects discrepancy (entering \emph{Dead reckoning} state for too long) while following the previous path. The intuition is that the location where robot detects the discrepancy is likely distinct. See Sec.~\ref{sec:resolve_ambiguity} for an example. In the worst case where such distinctive anchor is absent, robot might follow a cycle of anchors without making progress. The solution is to count how many times the robot has visited an anchor (i.e., by collecting statistics from  \emph{last visited anchor}). Cyclic behavior can be detected so that the robot can break the loop by biasing its choice in future planning. We leave the handling of this extreme case as future work.

\subsection{Following a Trajectory}
\label{sec:follow_traj}
Our trajectory follower constantly updates and tracks an active anchor to make progress, while performing dead reckoning to counter local disturbances. Specifically, given a sequence of anchor observations $\hat{o}_1, \hat{o}_2, ..., \hat{o}_m$, the trajectory follower acts as a state machine:

{\bf Search}: robot searches for the best anchor: $\hat{o}^* = \argmax_{o\in\{\hat{o}_1, ..., \hat{o}_m\}}\RE(o_{\text{current}}, o)$. If $\RE(o_{\text{current}}, \hat{o}^*) > p_{\text{search}}$, it sets $\hat{o}^*$ as current active anchor $\hat{o}_{\text{active}}$ and enters \emph{Follow} state, otherwise it gives up and stops.

{\bf Follow}: robot computes the next waypoint $x, y = \C_{\text{WP}}(o_{\text{current}}, \hat{o}_{\text{active}})$ and uses it to drive $\C_{\text{RMP}}$. Meanwhile it tracks and updates the following two values:
\begin{itemize}
    \item \emph{last visited anchor}. Robot uses the predicted mutual image overlap to measure the proximity between $o_{\text{current}}$ and anchors close to $\hat{o}_{\text{active}}$. The closest anchor is set as $o_{\text{lastvisited }}$. This is a form of approximate localization. 
    \item \emph{active anchor}. If $\RE(o_{\text{current}}, \hat{o}_{\text{active+1}}) > p_{\text{follow}}$ and is within proximity, it advances $\hat{o}_{\text{active}}$ to $\hat{o}_{\text{active + 1}}$, otherwise $\hat{o}_{\text{active}} = o_{\text{lastvisited + 1}}$. The intuition is to choose an $\hat{o}_{\text{active}}$ that is neither too close nor too far away.
\end{itemize}
Normally robot stays in \emph{Follow} state. But if moving obstacles or actuation noise cause $\RE(o_{\text{current}}, \hat{o}_{\text{active}}) < p_{\text{follow}}$, it enters \emph{Dead reckoning} state.

{\bf Dead reckoning}: robot tracks the last waypoint computed in the \emph{Follow} state and uses the waypoint to drive $\C_{\text{RMP}}$. The assumption is that disturbances are transient which the robot could escape by following the last successfully computed waypoint. Waypoint tracking can be done by an odometer and needs not be very accurate. While in this state, robot keeps checking if $\RE(o_{\text{current}}, \hat{o}_{\text{active}}) >  p_{\text{follow}}$ and returns to  \emph{Follow} state if possible.


\section{Experiments}
\label{sec:experiments}
We trained $\C_{\text{WP}}$, $\RE$ and all baselines in 12 Gibson environments. 100k training trajectories were generated by running an $\text{A}^*$ planner (used to provide waypoints) with a laser RMP controller similar to \cite{meng2019neural}. Simulation step size is $0.1$. We use the laser RMP controller as $\C_{\text{RMP}}$ mostly for efficiency, but in practice an image-based RMP controller can also be used \cite{meng2019neural}. $\C_{\text{WP}}$ was trained by randomly sampling two images on the same trajectory with certain visual overlap, with the $\text{A}^*$ waypoint as supervision. After $\C_{\text{WP}}$ was trained, we trained $\RE$ by sampling two images that either belong to the same trajectory (prob 0.6) or different trajectories (prob 0.4), and ran a rollout with $\C$ to get a binary outcome. Image size is $64\times 64$ with $120^\circ$ horizontal field of view. We augmented the dataset by jittering robot's starting location and orientation to improve generalization. About 1.5M samples were used to train $\C_{\text{WP}}$ and $\RE$. Our training setup models a real vehicle similar to \cite{racecar}, so that the same model can be used for real experiments.

We present quantitative results in 5 unseen Gibson environments with diverse appearances. Our baseline is based on SPTM. Since SPTM is designed for small synthetic mazes with discrete action space, its original version would perform poorly in our setting. For a fair comparison, we let SPTM use the same controller and trajectory following logic as ours. The main differences between SPTM and ours are thus: i) how reachability is learned and ii) how graph is constructed and used. Our ablation study will thus be in the form of evaluating trajectory following and planning performance in the following sections.

\subsection{Trajectory Sparsification}
\begin{figure}
    \begin{minipage}[b]{0.5\columnwidth}
    \centering
    \includegraphics[width=0.95\textwidth]{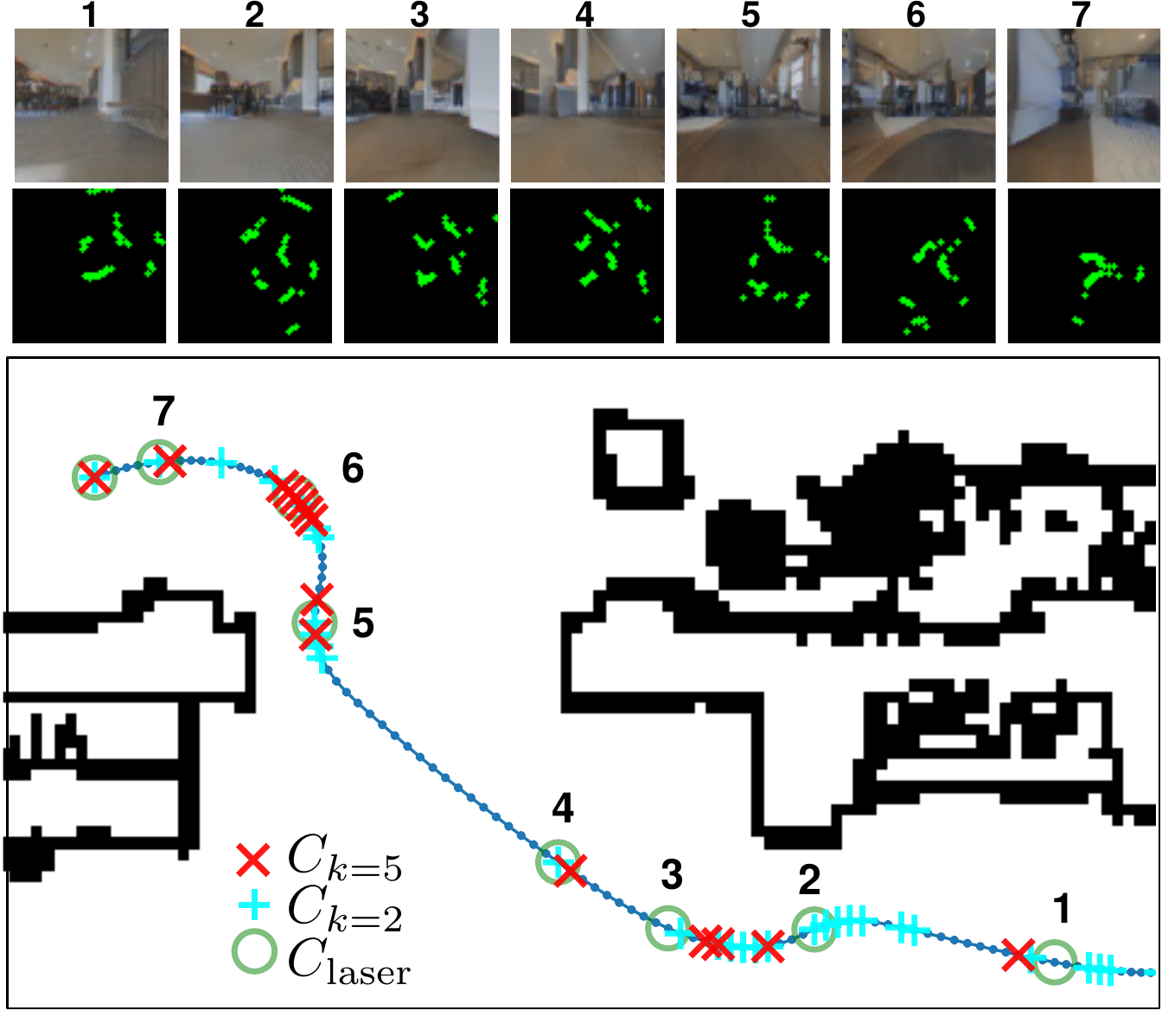}
    \caption{Trajectory sparsification. Blue dots: dense observations. Images correspond to numbered locations. 1D laser scans are visualized from the top view.}
    \label{fig:sparsification}    
    \end{minipage}\hfill\begin{minipage}[b]{0.47\columnwidth}
    \centering
    \hspace{2mm}\includegraphics[width=0.88\textwidth]{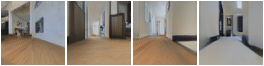}
    \vspace{0.5mm}
    \includegraphics[width=0.95\textwidth]{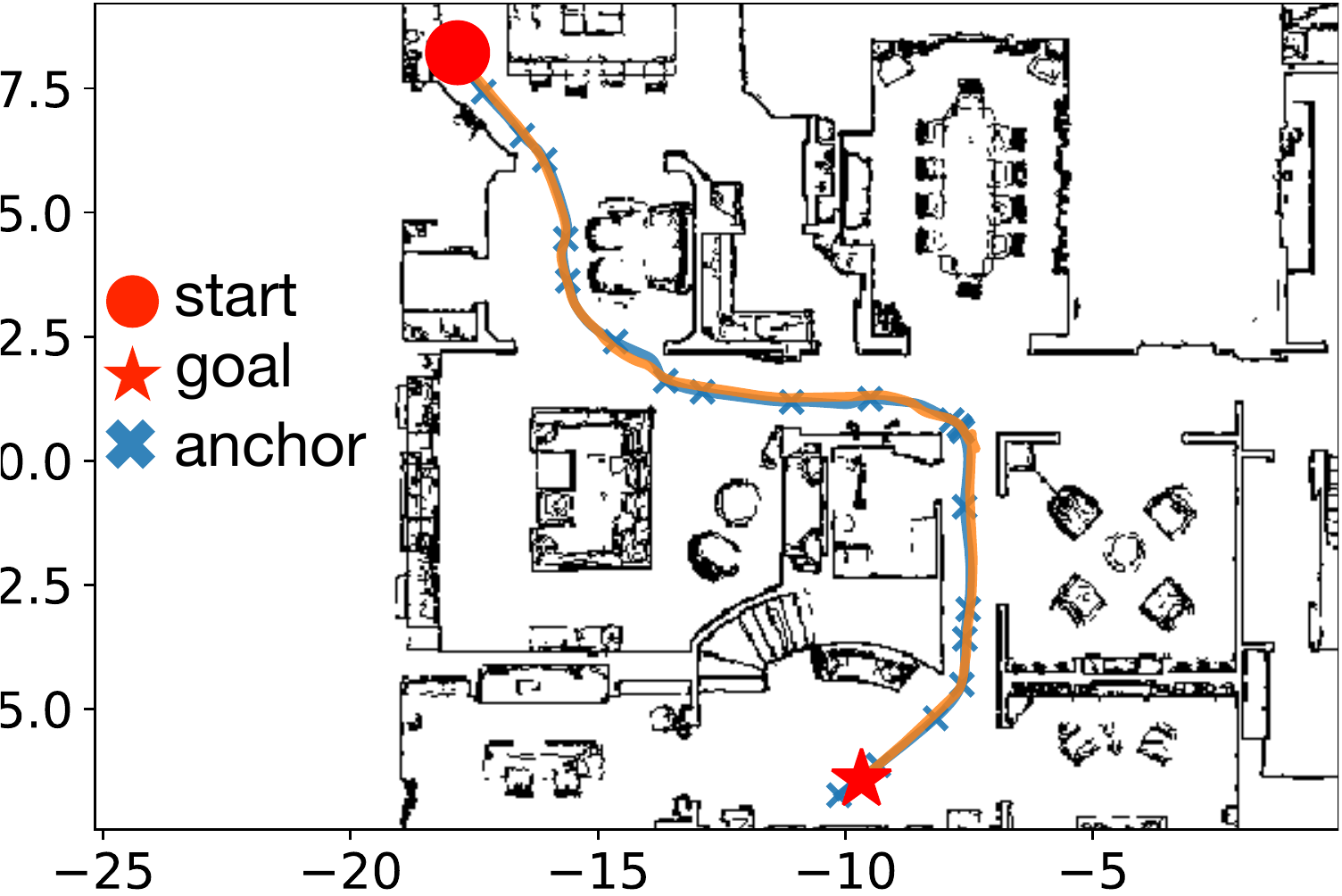}
    \caption{Following a 23m long trajectory. Blue trace: groundtruth trajectory. Orange trace is generated by following the anchor points.}
    \label{fig:traj_following_examples}
    \end{minipage}
    \figurespace
\end{figure}
Fig.~\ref{fig:sparsification} compares sparsification results of three controllers. The two visual controllers $\C_{k=2}$, $\C_{k=5}$ differ in their context length. To show that our model is general, we also trained a laser-based controller $\C_{\text{laser}}$ by modifying the input layer in Fig.~\ref{fig:network} to take 64-point $240^\circ$ 1D depth as input. 

Fig.~\ref{fig:sparsification} shows placement of anchors with $p_{\text{sparsify}}=0.99$. Comparing with $\C_{k=5}$, $\C_{k=2}$ requires denser anchors. Since $\C_{k=2}$ uses a shorter context, it is more ``local'' and has to keep more anchors to follow a path robustly. Nonetheless, anchors are more densely distributed in tight spaces and corners for both controllers, indicating that our sparsification strategy adapts well to environmental geometry. Interestingly, $\C_{\text{laser}}$ shows a more uniform distribution pattern. Since laser scans have a much wider field of view and measures geometry directly, it is not heavily affected by tight spaces and large viewpoint change.


\subsection{Trajectory Following}
\label{sec:traj_following}
We randomly generated 500 trajectories in the test environments (Fig.~\ref{fig:traj_following_examples}) with an average length of 15~m. When following a trajectory, we stop the robot when it diverges from the path or collides with obstacles. We report the cover rate, the percentage of total length of trajectories successfully followed by robot. For our trajectory followers, $p_{\text{search}}=p_{\text{follow}}=0.92$.

\begin{figure}
    \centering
    \includegraphics[width=0.8\columnwidth]{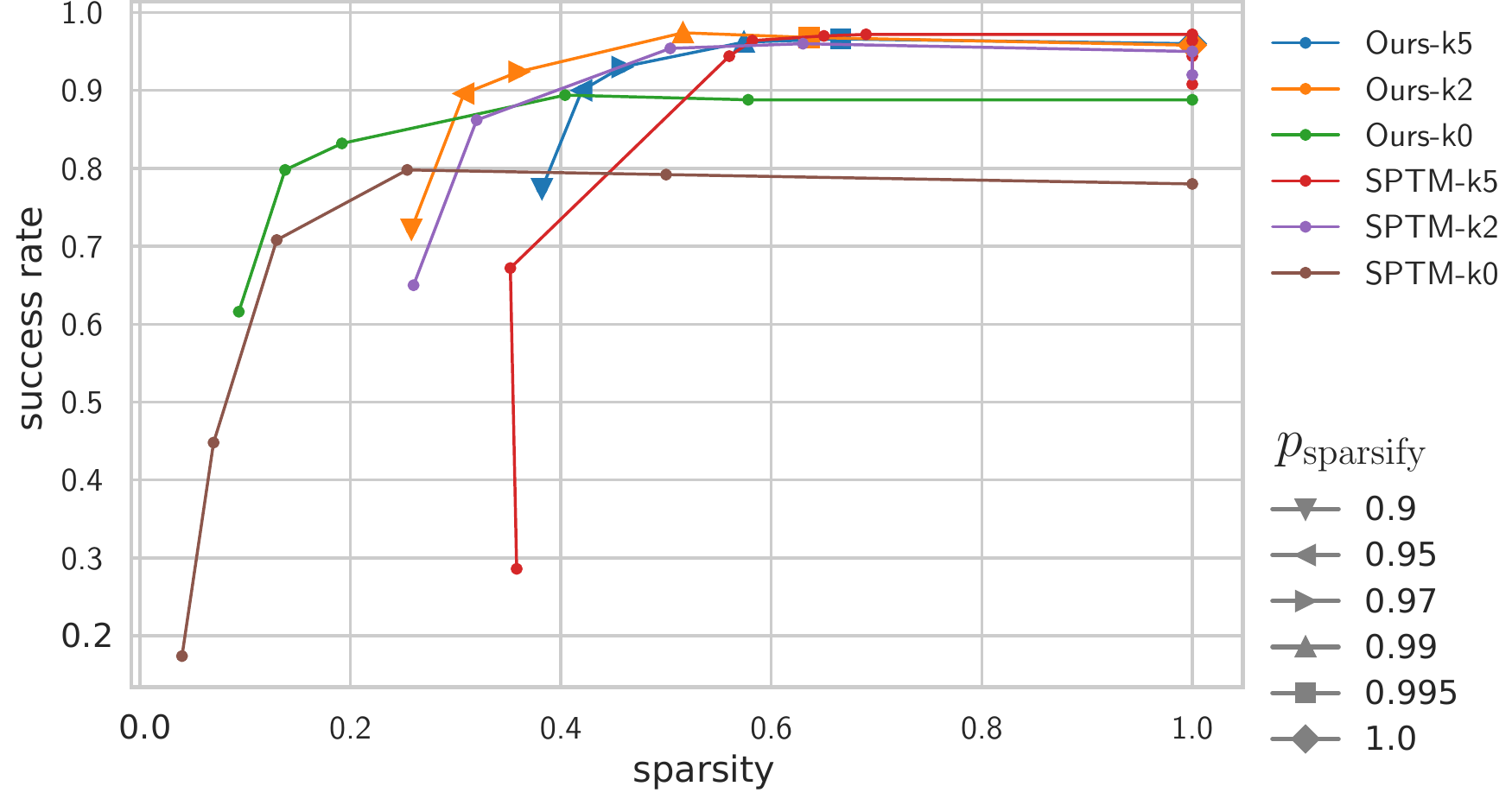}
    \caption{Trajectory following cover rate in 5 test environments. Number after $k$ indicates the context length. Data points for Ours-k5 and Ours-k2 are marked with $p_{\text{sparsify}}$.}
    \label{fig:traj_follow_success_rate}
    \figurespace
    \vspace{2mm}
\end{figure}

Sparsity is the average ratio of number of images in a sparsified trajectory to the number of images in the corresponding dense trajectory. To change sparsity, we vary $p_\text{sparsify}$ for our models. For SPTM we select every $n$th frame and vary $n$. Fig.~\ref{fig:traj_follow_success_rate} plots cover rates for varying sparsity conditions. Controllers with contexts (*-k2, *-k5) achieve higher than 95\% cover rate, better than controllers without context (at most 90\%). This indicates that having contextual frames can improve robustness. But since contextual frames are used, more observations have to be kept so storage-wise it is not as efficient as (*-k0). 


SPTM performs comparably to ours when using a strong controller (*-k5), but for all controllers it starts to degrade before ours as sparsity lowers. Due to its fixed-interval subsampling, it does not adapt to controllers' capability well, as can be seen by the increasing gap between ours and the SPTM counterparts when less contextual frames are used (*-k2, *-k0).

We also evaluated performance under noisy actuation by multiplying a random scaling factor $s\sim\mathcal{N}(1.0, 0.33)$ to the control output. No noticeable difference was found. This is expected because the local controller runs at a high frequency (10~Hz) and uses visual feedback for closed-loop control.


\subsection{Planning}
\subsubsection{Navigation between Places}
\label{sec:nav_between_places}
\begin{figure}
    \captionsetup[subfigure]{aboveskip=0pt,belowskip=-1pt}
    \centering
    \begin{subfigure}[b]{0.49\columnwidth}
    \centering
    \includegraphics[width=1\textwidth]{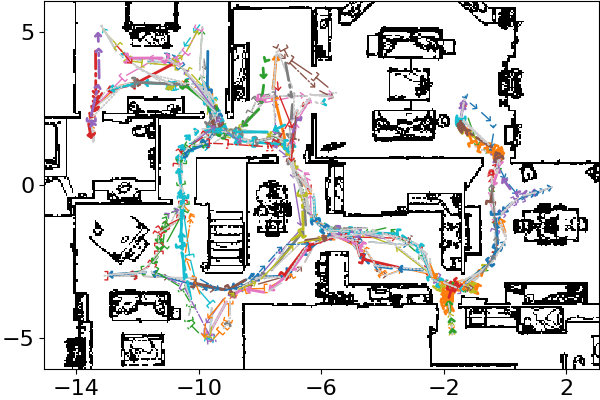}
    \caption{Gibson house24}
    \end{subfigure}
    \begin{subfigure}[b]{0.49\columnwidth}
    \centering
    \includegraphics[width=1\textwidth]{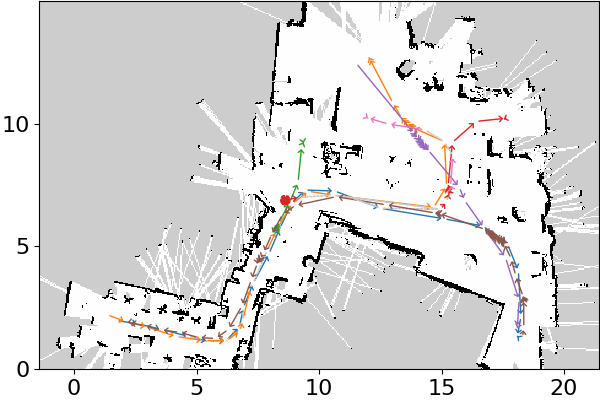}
    \caption{A real environment}
    \end{subfigure}
    \caption{Example topological maps built from sparsified trajectories. Each trajectory is assigned a different color.}
    \label{fig:nav_graph}
    \figurespace
    \vspace{-0.5ex}
\end{figure}

We built one topological map for each environment (Fig.~\ref{fig:nav_graph}a). A map is constructed from 90 trajectories connecting 10 locations in a pairwise fashion. The locations are selected to make the trajectories cover most of the reachable area.

Robot starts at one of the locations (with jittered position and orientation) and is given an goal image taken at one of the other 9 destinations. Robot has no prior knowledge of its initial location. We re-implemented SPTM's planner and uses the best trajectory follower SPTM-k5 (Sec\ref{sec:traj_following}) to make it a competitive baseline. We set the sub-sampling ratio to 20 and $\Delta T_l=1$ to prevent the graph from getting too large. We performed a hyperparameter search to set $s_{\text{shortcut}}=0.99$. 
For our method, we set $p_{\text{reuse}}=p_{\text{edge}}=0.99$.


Table~\ref{tab:planning_success_rate} presents the success rate for each environment compared to SPTM. Our method outperforms SPTM with much sparser maps. Graphs built by SPTM have unweighted edges and do not reuse vertices. SPTM also does explicit localization which sometimes causes planning failure. This results in worse scalability and reliability compared with our approach. Note that the slightly lower success rates in \textit{space8} and \textit{house75} are mostly caused by strong symmetry and rendering artefacts.  



\begin{table}
  \centering
  \footnotesize
  \setlength{\tabcolsep}{3pt}
  \begin{tabular}{l|*{5}{c}}
    \toprule
    & space8
    & house24
    & house29
    & house75
    & house79\\
    \hline
    Area
    & $460m^2$
    & $207m^2$
    & $270m^2$
    & $403m^2$
    & $205m^2$\\
    Images
    & 30,342 & 31,167 & 28,679 & 39,788 & 33,617\\[2pt]
    \hline
    SPTM & 1,648/3,201 & 1,688/3,668 & 1,560/3,960 & 2,116/4,115 & 1,808/4,756\\[2pt]
         & 48.1\% & 40.2\% & 45.6\% & 51.3\% & 47.2\%\\[2pt]
    \hline
    Ours & 974/1,482 & 900/1,348 & 901/1,467 & 1,454/2,275 & 909/1,524\\[2pt]
         & 86.9\% & 94.3\% & 91.2\% & 84.6\% & 95.7\%\\
    \bottomrule
  \end{tabular}
  \caption{Planning success rate, with \#vertices/\#edges shown above. Success rate is the outcome of 1,000 navigation trials. }
  \label{tab:planning_success_rate}
  \figurespace
\end{table}

\subsubsection{Comparing Trajectory Probabilities to Empirical Success Rate}
\begin{figure}[t]
\begin{minipage}[t]{0.57\columnwidth}
    \centering
    \hspace{-5mm}\includegraphics[scale=0.48]{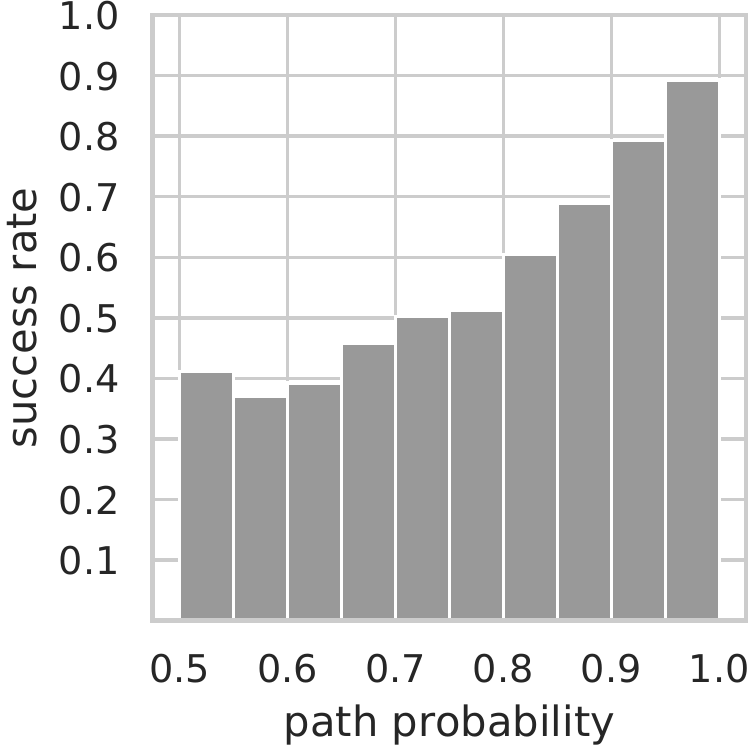}
    \caption{Comparing path probability to empirical success rate. Each bar is the average success rate of paths whose probabilities fall into that range.}
    \label{fig:path_prob_success_rate}
\end{minipage}\hfill\begin{minipage}[t]{0.39\columnwidth}
    \centering
    \hspace{-10mm}\includegraphics[scale=0.48]{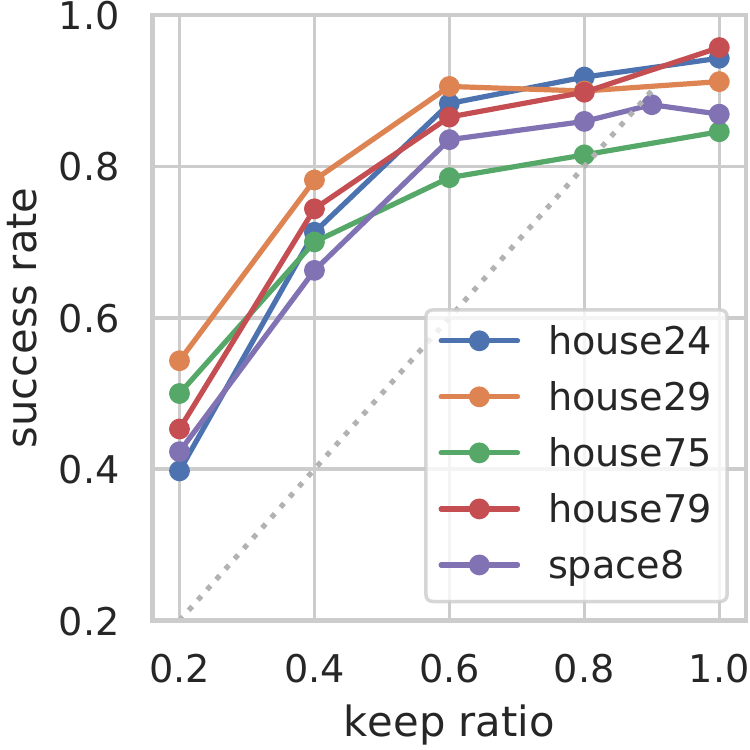}
    \caption{Success rate with pruned graphs. Dotted line shows no-generalization for reference.}
    \label{fig:subgraph_success_rate}
\end{minipage}
\figurespace
\end{figure}

To show that path probability is a reasonable indicator of empirical outcome, we let a robot start at a random location (anywhere in a map), plan a path to one of the 10 destinations, and follow the path. 1,000 trajectories were collected in each environment. Fig.~\ref{fig:subgraph_success_rate} shows that path probability strongly corresponds to empirical success rate. This allows a robot to assess the risk before executing a plan, and ask for help if necessary. Note that SPTM does not provide any uncertainty measure.

\subsubsection{Generalizing to New Navigation Tasks}
\begin{figure}[t]
\begin{minipage}[t]{0.34\columnwidth}
    \centering
    \includegraphics[width=0.95\textwidth]{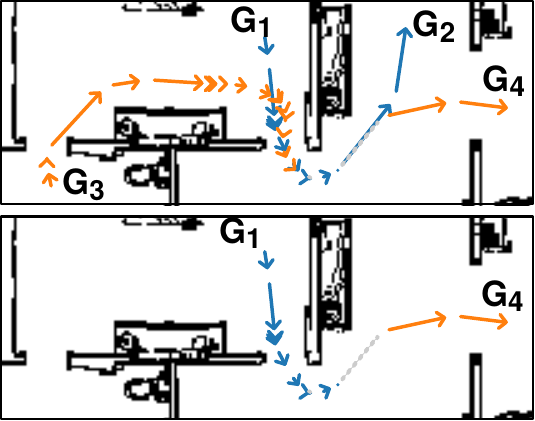}
    \caption{Plans a path from $G_1$ to $G_4$ by combining $G_1\rightarrow G_2$ and $G_3\rightarrow G_4$}
    \label{fig:path_generalization}
\end{minipage}\hfill\begin{minipage}[t]{0.62\columnwidth}
    \centering
    \includegraphics[width=0.95\textwidth]{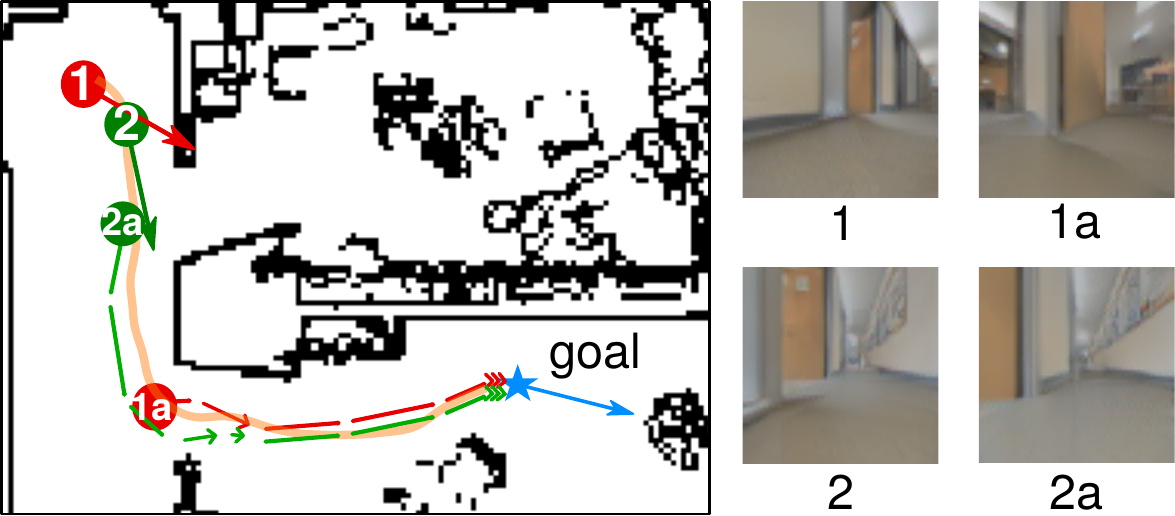}
    \caption{Online planning. Arrows indicate headings. 1a and 2a are the next anchor observations for the two paths respectively.}
    \label{fig:online_planning}
\end{minipage}
\figurespace
\end{figure}

To test the generalizability of our planner, we randomly pruned the graphs to contain only a subset of the trajectories, and repeated the experiment in \ref{sec:nav_between_places}. Fig.~\ref{fig:subgraph_success_rate} shows that with only 60\% of the trajectories, robot already performs close to its peak success rate. In other words, robot is able to combine existing trajectories to solve novel navigation tasks. Fig.~\ref{fig:path_generalization} shows an example. 

\subsubsection{Resolving Ambiguity}
\label{sec:resolve_ambiguity}
Fig.~\ref{fig:online_planning} illustrates how perceptual aliasing is resolved in environments with strong symmetry. Robot initially starts at an ambiguous location (marked ``1'') and plans a wrong path (red path). While following this path, robot detects the discrepancy at ``2'' by realizing what is expected to be an office room is actually a hallway. As a result, it plans a new path (green) whereby it successfully reaches the goal.


\subsubsection{Scalability}
\begingroup
\setlength{\columnsep}{5pt}
\begin{wrapfigure}{L}{0.4\columnwidth}
    \vspace{-3ex}
    \centering
    \includegraphics[width=0.4\columnwidth]{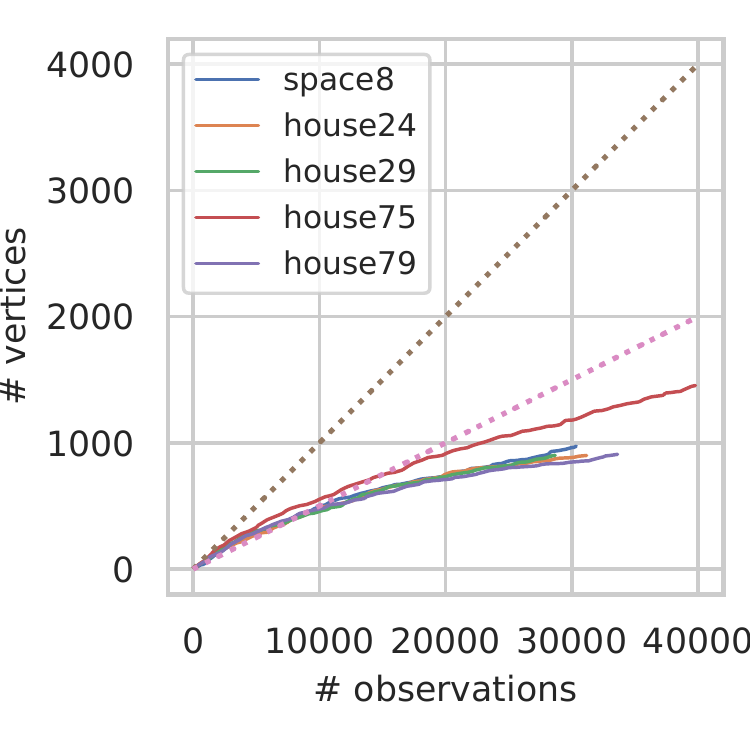}
    \caption{Dotted lines show 10x and 20x temporal subsampling for reference.}
    \label{fig:scalability}
\end{wrapfigure}
Fig.~\ref{fig:scalability} shows that map sizes grow sublinearly to the number of raw observations, making our approach  scalable to large environments. It also shows that map size is adaptive to an environment. Since house75 has more complex geometry and exhibits more rendering artefacts, denser samples are kept to stay conservative.

\endgroup

\subsection{Testing in a Real Environment}

Our model trained in simulation generalizes well to real images without finetuning. To map a real environment, we manually drove the RC car to collect 7 trajectories, totalling 3,200 images. The final map contains 206 vertices and 215 edges (Fig.\ref{fig:nav_graph}b). The car is able to plan novel paths between locations and follow the path while avoiding obstacles not seen during mapping (Fig.\ref{fig:intro}). We refer the interested reader to the video supplementary material for more examples.

\section{Conclusion}
In this work, we show that by learning the capability of a local controller, robust and scalable visual topological navigation can be achieved. Due to the simplicity and flexibility of our framework, it can be extended to support non-visual sensors and applied to other robotics problems. Future works include combining multiple sensors to improve the controller, developing better algorithms to resolve ambiguity, improving generalization, and extending to manipulation tasks.

The hyperparameters in our approach are mostly probability thresholds, which are easy to interpret and tune. One important scenario our approach does not handle is when robot deviates too much from all vertices in the navigation graph, where it would fail to find a plausible path. A self-exploratory model can help here, and it can also be used for autonomous map construction. 

\section{Acknowledgements}
This work was funded in part by ONR grant 63-6094 and by the Honda Curious Minded Sponsored Research Agreement. We thank NVIDIA for generously providing a DGX used for this research via the NVIDIA Robotics Lab and the UW NVIDIA AI Lab (NVAIL).

\bibliographystyle{IEEEtran}
\bibliography{IEEEabrv,references.bib}

\begin{thebibliography}{10}
\providecommand{\url}[1]{#1}
\csname url@rmstyle\endcsname
\providecommand{\newblock}{\relax}
\providecommand{\bibinfo}[2]{#2}
\providecommand\BIBentrySTDinterwordspacing{\spaceskip=0pt\relax}
\providecommand\BIBentryALTinterwordstretchfactor{4}
\providecommand\BIBentryALTinterwordspacing{\spaceskip=\fontdimen2\font plus
\BIBentryALTinterwordstretchfactor\fontdimen3\font minus
  \fontdimen4\font\relax}
\providecommand\BIBforeignlanguage[2]{{%
\expandafter\ifx\csname l@#1\endcsname\relax
\typeout{** WARNING: IEEEtran.bst: No hyphenation pattern has been}%
\typeout{** loaded for the language `#1'. Using the pattern for}%
\typeout{** the default language instead.}%
\else
\language=\csname l@#1\endcsname
\fi
#2}}

\bibitem{savinov2018semi}
N.~Savinov, A.~Dosovitskiy, and V.~Koltun, ``Semi-parametric topological memory
  for navigation,'' in \emph{International Conference on Learning
  Representations ({ICLR})}, 2018.

\bibitem{wubayesian}
Y.~Wu, Y.~Wu, A.~Tamar, S.~Russell, G.~Gkioxari, and Y.~Tian, ``Bayesian
  relational memory for semantic visual navigation,'' in \emph{International
  Conference on Computer Vision (ICCV)}, 2019.

\bibitem{kumar2018visual}
A.~Kumar, S.~Gupta, D.~Fouhey, S.~Levine, and J.~Malik, ``Visual memory for
  robust path following,'' in \emph{Advances in Neural Information Processing
  Systems}, 2018, pp. 765--774.

\bibitem{blochliger2018topomap}
F.~Blochliger, M.~Fehr, M.~Dymczyk, T.~Schneider, and R.~Siegwart, ``Topomap:
  Topological mapping and navigation based on visual slam maps,'' in \emph{IEEE
  International Conference on Robotics and Automation (ICRA)}, 2018.

\bibitem{chen2019graphnav}
K.~Chen, J.~P. de~Vicente, G.~Sepulveda, F.~Xia, A.~Soto, M.~Vazquez, and
  S.~Savarese, ``A behavioral approach to visual navigation with graph
  localization networks,'' \emph{Robotics Science and Systems (RSS)}, 2019.

\bibitem{thrun2005probabilistic}
S.~Thrun, W.~Burgard, and D.~Fox, \emph{Probabilistic robotics}.\hskip 1em plus
  0.5em minus 0.4em\relax MIT press, 2005.

\bibitem{mur2015orb}
R.~Mur-Artal, J.~M.~M. Montiel, and J.~D. Tardos, ``Orb-slam: a versatile and
  accurate monocular slam system,'' \emph{IEEE Transactions on Robotics},
  vol.~31, no.~5, pp. 1147--1163, 2015.

\bibitem{hirose2019deep}
N.~{Hirose}, F.~{Xia}, R.~{Martín-Martín}, A.~{Sadeghian}, and S.~{Savarese},
  ``Deep visual mpc-policy learning for navigation,'' \emph{IEEE Robotics and
  Automation Letters}, vol.~4, no.~4, pp. 3184--3191, 2019.

\bibitem{rmp}
N.~D. Ratliff, J.~Issac, and D.~Kappler, ``Riemannian motion policies,''
  \emph{CoRR}, vol. abs/1801.02854, 2018.

\bibitem{xia2018gibson}
F.~Xia, A.~R. Zamir, Z.~He, A.~Sax, J.~Malik, and S.~Savarese, ``Gibson env:
  Real-world perception for embodied agents,'' in \emph{IEEE International
  Conference on Computer Vision and Pattern Recognition (CVPR)}, 2018, pp.
  9068--9079.

\bibitem{o1978hippocampus}
J.~O'keefe and L.~Nadel, \emph{The hippocampus as a cognitive map}.\hskip 1em
  plus 0.5em minus 0.4em\relax Oxford: Clarendon Press, 1978.

\bibitem{hafting2005microstructure}
T.~Hafting, M.~Fyhn, S.~Molden, M.-B. Moser, and E.~I. Moser, ``Microstructure
  of a spatial map in the entorhinal cortex,'' \emph{Nature}, vol. 436, no.
  7052, p. 801, 2005.

\bibitem{doeller2010evidence}
C.~F. Doeller, C.~Barry, and N.~Burgess, ``Evidence for grid cells in a human
  memory network,'' \emph{Nature}, vol. 463, no. 7281, p. 657, 2010.

\bibitem{thrun1998learning}
S.~Thrun, ``Learning metric-topological maps for indoor mobile robot
  navigation,'' \emph{Artificial Intelligence}, vol.~99, no.~1, pp. 21--71,
  1998.

\bibitem{milford2004ratslam}
M.~J. Milford, G.~F. Wyeth, and D.~Prasser, ``Ratslam: a hippocampal model for
  simultaneous localization and mapping,'' in \emph{IEEE International
  Conference on Robotics and Automation (ICRA)}, vol.~1, 2004, pp. 403--408.

\bibitem{kuipers2000spatial}
B.~Kuipers, ``The spatial semantic hierarchy,'' \emph{Artificial intelligence},
  vol. 119, no. 1-2, pp. 191--233, 2000.

\bibitem{maddern2012capping}
W.~Maddern, M.~Milford, and G.~Wyeth, ``Capping computation time and storage
  requirements for appearance-based localization with cat-slam,'' in \emph{2012
  IEEE International Conference on Robotics and Automation (ICRA)}, 2012, pp.
  822--827.

\bibitem{fraundorfer2007topological}
F.~Fraundorfer, C.~Engels, and D.~Nist{\'e}r, ``Topological mapping,
  localization and navigation using image collections,'' in \emph{International
  Conference on Intelligent Robots and Systems (IROS)}, 2007.

\bibitem{cummins2011appearance}
M.~Cummins and P.~Newman, ``Appearance-only slam at large scale with fab-map
  2.0,'' \emph{The International Journal of Robotics Research}, 2011.

\bibitem{linegar2015work}
C.~Linegar, W.~Churchill, and P.~Newman, ``Work smart, not hard: Recalling
  relevant experiences for vast-scale but time-constrained localisation,'' in
  \emph{IEEE International Conference on Robotics and Automation (ICRA)}.\hskip
  1em plus 0.5em minus 0.4em\relax IEEE, 2015, pp. 90--97.

\bibitem{Savinov2019_EC}
N.~Savinov, A.~Raichuk, R.~Marinier, D.~Vincent, M.~Pollefeys, T.~Lillicrap,
  and S.~Gelly, ``Episodic curiosity through reachability,'' in
  \emph{International Conference on Learning Representations ({ICLR})}, 2019.

\bibitem{bansal2019combining}
S.~Bansal, V.~Tolani, S.~Gupta, J.~Malik, and C.~Tomlin, ``Combining optimal
  control and learning for visual navigation in novel environments,''
  \emph{Conference on Robot Learning (CoRL)}, 2019.

\bibitem{pathak2018zero}
D.~Pathak, P.~Mahmoudieh, G.~Luo, P.~Agrawal, D.~Chen, Y.~Shentu, E.~Shelhamer,
  J.~Malik, A.~A. Efros, and T.~Darrell, ``Zero-shot visual imitation,'' in
  \emph{International Conference on Learning Representations ({ICLR})}, 2018.

\bibitem{gupta2017cognitive}
S.~Gupta, J.~Davidson, S.~Levine, R.~Sukthankar, and J.~Malik, ``Cognitive
  mapping and planning for visual navigation,'' in \emph{IEEE International
  Conference on Computer Vision and Pattern Recognition (CVPR)}, 2017, pp.
  7272--7281.

\bibitem{gupta2017unifying}
S.~Gupta, D.~Fouhey, S.~Levine, and J.~Malik, ``Unifying map and landmark based
  representations for visual navigation,'' \emph{arXiv preprint
  arXiv:1712.08125}, 2017.

\bibitem{mirowski2018learning}
P.~Mirowski, M.~Grimes, M.~Malinowski, K.~M. Hermann, K.~Anderson,
  D.~Teplyashin, K.~Simonyan, A.~Zisserman, R.~Hadsell, \emph{et~al.},
  ``Learning to navigate in cities without a map,'' in \emph{Advances in Neural
  Information Processing Systems}, 2018, pp. 2419--2430.

\bibitem{mirowski2016learning}
P.~Mirowski, R.~Pascanu, F.~Viola, H.~Soyer, A.~J. Ballard, A.~Banino,
  M.~Denil, R.~Goroshin, L.~Sifre, K.~Kavukcuoglu, \emph{et~al.}, ``Learning to
  navigate in complex environments,'' \emph{International Conference on
  Learning Representations ({ICLR})}, 2018.

\bibitem{tedrake2010lqr}
R.~Tedrake, I.~R. Manchester, M.~Tobenkin, and J.~W. Roberts, ``Lqr-trees:
  Feedback motion planning via sums-of-squares verification,'' \emph{The
  International Journal of Robotics Research}, vol.~29, no.~8, pp. 1038--1052,
  2010.

\bibitem{meng2019neural}
X.~Meng, N.~Ratliff, Y.~Xiang, and D.~Fox, ``Neural autonomous navigation with
  riemannian motion policy,'' in \emph{IEEE International Conference on
  Robotics and Automation (ICRA)}, 2019.

\bibitem{racecar}
\BIBentryALTinterwordspacing
``{MIT racecar},'' 2018. [Online]. Available:
  \url{https://mit-racecar.github.io/}
\BIBentrySTDinterwordspacing

\end{thebibliography}

\end{document}